# 5. Technological Approaches to Detecting Online Disinformation and Manipulation

*Aleš Horák, Vít Baisa, Ondřej Herman*

## 5.1 Introduction

The move of propaganda and disinformation to the online environment is possible thanks to the fact that within the last decade, digital information channels radically increased in popularity as a news source. The main advantage of such media lies in the speed of information creation and dissemination. This, on the other hand, inevitably adds pressure, accelerating editorial work, fact-checking, and the scrutiny of source credibility.

In this chapter, an overview of computer-supported approaches to detecting disinformation and manipulative techniques based on several criteria is presented. We concentrate on the technical aspects of automatic methods which support fact-checking, topic identification, text style analysis, or message filtering on social media channels. Most of the techniques employ artificial intelligence and machine learning with feature extraction combining available information resources. The following text firstly specifies the tasks related to computer detection of manipulation and disinformation spreading. The second section presents concrete methods of solving the tasks of the analysis, and the third sections enlists current verification and benchmarking datasets published and used in this area for evaluation and comparison.

## 5.2 Task specification

With the growth of digital-born and social media publishing, the origin of news distributed to the general public can be easily performed and the process of publishing a message to a wide group of readers is extremely simplified. This opens new possibilities for any pressure group to publish disinformation or purpose-modified news, which are expressed so as to be apparently accepted as objective reporting of current events (Woolley and Howard 2018). On the other hand, the predetermined availability of the texts in an online digital form opens new possibilities for the detection of such persuasive techniques through the analysis of the content, the style of the text, and its broader context.

The presented approaches usually distinguish the two main aspects of the analysed texts: (a) whether the text is intentionally truthless (disinformation, fake news), or (b) whether the text refers to an actual event or situation but the form and content is adapted from an objective description for a reason (propaganda, manipulation). Note that the first category usually does not include 'misinformation', that is, texts which are unintentionally truthless, where their author is convinced the message is faithful (see chapter 1).

### 5.2.1 Fake news detection

In general, a fully capable technical method to recognise that an input text contains truthless statement(s) would have to be implemented as an omniscient oracle. In practice, the task of fake news detection uses various second-level aspects of the text, which can be handled by a thorough analysis of the available data (Kavanagh, Rich 2018). Fake news detection approaches include:

- fact-checking,
- source credibility analysis,
- information flow analysis, and
- manipulative style recognition.

The *fact-checking* or *verification* approach implements the most logical idea of verifying whether a message is true or not, which resembles the same process executed by a human expert—a journalist when processing the input information, for example. In the automated approach, the text first needs to be analysed using natural language processing (NLP) methods, and individual objective fact statements or claims are identified. Consequently, each individual fact is verified—confirmed or refuted, usually with a confidence score—by comparing the statement to a predefined source of knowledge, for example, knowledge graphs built from Wikipedia texts (Ciampaglia et al. 2015); manual datasets of verified facts prepared by human experts, such as CNN Facts First or PolitiFact by the Poynter Institute (Hassan et al. 2017); or collaborative verification by a network of engaged specialists, such as CaptainFact or CrossCheck (Cazalens et al. 2018). Collaborative verification can be used in the broader sense of the claim verification process as the involved community can also judge the truthfulness of images and video content, which is still unrealistic with automated tools. In the process of simplifying and distributing the process of fact annotation, Duke University and Google Research (Adair et al. 2017) have published an open standard entitled ClaimReview, which details how to annotate the result of checking someone else's claims. Accessible online ClaimReviews are then aggregated and used for the continuous expansion of fact-checking databases. Since the ratio of possible errors in the fact processing chain is still rather high, practical tools for fact-checking usually offer multiple fact verification sources and offer the automated results as expert supporting tools where the final decision is left to the human specialist. Exploiting fact-supporting tools became such an important part of serious news preparation that a completely new style of editing has been defined as 'computational' or 'digital journalism' (Caswell, Anderson 2019).

*Source credibility analysis* exploits information about the origin of the text. This method can of course be inclined to counterfeiting or interpolating the source identification, but in cases where the text author is sufficiently documented, source credibility offers an important piece of information. The simplest case of source analysis takes the form of consulting a trustworthy database of internet IP addresses as this is the primary information about where an online text originated. An example of such a general database (and accompanying tools, e.g. web browser extensions) is the Adblock Plus tool, which allows for the management of multiple blacklisted (forbidden) and whitelisted (allowed) sources, including external community created lists, such as EasyList (Wills, Uzunoglu 2016). Intensive efforts in source certification widen the simple address-based judgement with transparency rules for best practices, covering citation and references work, reporter expertise, or other trust indicators. Even though those approaches can include analysis via automated tools, the main aspect of the credibility decisions is driven by human experts organised in an established initiative, such as Trust Project Indicators or the International Fact-Checking Network (Graves 2016).

According to Shearer (2018), more than two-thirds of US adults receive news via social media such as Facebook, YouTube, or Twitter. These networks provide detailed information about the process of sharing each message across the network and thus open the possibility of an *information flow analysis*. For example, TwitterTrails (Finn et al. 2014) uses knowledge from the Twitter network about the originator, the burst time and the whole timeline of the message, the propagators, and other actors related to the message (including the audience), and it checks whether there are any refutations of the message in the flow. Vosoughi et al. (2018) have shown that false news spreads substantially faster and reaches more people than true stories; while the most retweeted true news arrived to about a thousand people, the same category of false news found their way to tens of thousands of readers. The Hoaxy Twitter analysis system by Indiana University (Shao et al. 2016) has gathered and published the Misinformation and Fact-Checking Diffusion Network, consisting of more than 20 million retweets with timestamp and user identifiers, allowing for the observation and quantification of all message flows between Twitter users, including important statistical information about user activity and URL popularity. The data allows automated bot users to be distinguished from real humans as well as the identification of influence centres or the origins of selected fake news.

The fourth category of fake news detection tasks leans on the fact that intentional disinformation is often expressed with a specific expressive style in order to promote the intended manipulation of the reader. *Manipulative style recognition* also belongs among the detection techniques described in the next section devoted to the recognition of propagandistic texts. With fake news, the characteristic style of the news usually forms a supplemental indicator and is analysed in combination with other factors. Volkova et al. (2017) designed a neural network model processing text, social graph and stylistic markers expressing bias, subjectivity, persuasiveness, and moral foundations cues. They showed that linguistic or stylistic analysis significantly improved the results.

### 5.2.2 Detection of manipulative techniques

Within this chapter, we define propagandistic news as texts which report about (at least partially) true events or facts but use specific *manipulative techniques*, framing the content of the messages in a purposeful way to promote political, economic, or other goals (see chapter 1). Basic algorithms to detect propaganda have been published for more than eighty years (Institute for Propaganda Analysis 1938), naturally for processing by people, not machines. The instructions presented seven stylistic techniques (propaganda devices) used by manipulators in the text, such as name-calling, transfer, or bandwagoning. In the current detection procedures, this main task remains very similar: to identify whether any specific reader manipulation technique exists in the text.[1]

Everybody has met at one point the most widespread example of simple propagandistic texts: spam email messages or web spam (Metaxas 2010). Unlike antispam techniques, which are based on the weighted occurrence scores of specific words and word combinations, the style of propaganda in news is analysed with complex metrics of *computational stylometry*. The question of whether the studied message contains a specific manipulative technique or not is here shifted to the question of whether the message is written in a specific *style* usually used with that manipulation device. Besides propaganda style detection, stylometric methods are used to recognise anonymous authorship attribution or personal information about the text author, such as his or her gender, age, education, or native language (Neal et al. 2017). The input data for stylometric methods is formed by a multitude of measurable quantitative features in the input text—besides the words and word combinations themselves, the metrics exploit information about the statistics of word and sentence lengths, world class co-occurrences, syntactic (sub)structures, emoticons, typographical and grammatical errors, punctuation marks, and so on. Within the algorithms of manipulative technique recognition, these features can be supplemented with information drawn from user profile analyses or publicly available traits of a previous user's behaviour, such as ratings or registration date (Peleschyshyn et al. 2016). In detailed analyses, especially when seeking specific explanations, the identification task orients from whole documents to individual sentences where a manipulative technique should be discovered (Da San Martino et al. 2019).

---

1 For detailed lists of the current analysed techniques, see section 4.4 'Datasets and evaluation'.

### 5.2.3 Generating fictive content

Language models are probabilistic devices which can predict the probability a sequence of words is a correct phrase in a language. Besides this function, language models may be also used for *generating artificial text*, which resembles a text written by a human. Until recently, generated texts were not able to 'fool' a human reader if the generated sequence was longer than one or two sentences. However, in 2019, the OpenAI group published a new neural model named GPT-2 (Radford et al. 2019) which was able to generate coherent newspaper articles of several paragraphs which sound authentic to a reader. The main reason for this change was the growth of both the data used for training as well as the size of the underlying neural network architecture. Following this, other current neural approaches, especially BERT (Devlin et al. 2019) and Grover (Zellers et al. 2019), proved it possible to generate thematically predetermined fictive news which is very difficult to distinguish from real human-generated newspaper texts. Zellers et al. (2019) showed that propaganda texts generated by Grover were on average evaluated as better in style than human written propaganda. In such a case, the opposite question of deciding whether a news text was written by a human or generated as a fictive one by a machine becomes crucial. Fortunately, the same techniques used for generations can also be exploited for detection, and, at this task, they reach super-human performance.

## 5.3 Methods specification

The amount of information posted online is very large and keeps growing as the Internet becomes more widely available and as users move from established media channels to consume and generate more content online. The immediate nature of the Internet combined with the humongous amount of content to be checked for malicious influence precludes any possibility of manual inspection of any significant part of online traffic before it is spread among more and more users. Various automated methods have been proposed to monitor and detect the actions of malicious actors online. In this section, we present a summary of these methods.

The methods can be broadly classified into four classes. (1) Fact-checking based methods inspect the information content of the articles. Automated knowledge extraction is still in its infancy, the usual approach is therefore semi-automatic, where the algorithmically extracted facts are verified by human annotators and then checked against a knowledge base. (2) Manipulative style recognition methods are based on the assumption that deception can be detected from surface features of the content. The cues include, for example, vocabulary used, sentiment, loaded language, subjectivity, and others. (3) Methods based on source credibility rely on the belief that unreliable users and sources have a higher probability of spreading deceptive information. Collaborative filtering based on crowdsourced data, where users vote on posts or articles, can be used. Based on these noisy votes, the aim is to extract a reliable signal from which posters or the posts themselves can be classified as malicious. A closely related research stream is based on (4) information flow analysis. The object of the study is the flow of information between different sources and users and the interaction between them.

As the task of fake news identification is only specified very vaguely, and no objective large-scale test set of fake news currently exists, comparison between different approaches is difficult. In the following sections, we describe the most common approaches to deception detection.

### 5.3.1 Fact-checking

Manual fact-checking by experts is a very reliable way of distinguishing fake news. However, it is very laborious and time-consuming and, therefore, expensive and not scalable with the amount of information being generated.

In addition to websites disseminating expert provided fact-checking, such as Snopes, which is one of the oldest websites debunking myths, or Hoax Slayer, dedicated mainly to combating email and social media hoaxes, websites aiming to provide crowdsourced fact-checking services have been appearing recently, for example, CrossCheck, Trive, or Fiskkit.

The fact-checking task can be split into two steps. The first step deals with the extraction of facts from the text, and, in the second step, the truthfulness of these facts is verified against a reliable knowledge base—in other words, a predefined database of (human) verified facts in a machine-readable format. A common knowledge representation format is the SPO (subject–predicate–object) triple as defined in the Resource Description Framework (Klyne et al. 2014). For example, the assertion 'The capital of France is Paris' would be represented as (Paris–capital_of–France). The knowledge can be understood as a knowledge graph, where the subjects and objects represent nodes and predicates form links between these nodes.

Approaches to knowledge base construction range from the manually built Freebase (Bollacker et al. 2008) or DBpedia (Auer et al. 2007), which extract structured facts from Wikipedia, to Knowledge Vault (Dong et al. 2014), which extracts facts from web content and also provides the probability of correctness of stored facts. These resources are mainly focused on common knowledge about the world, which changes relatively slowly, whereas fact-checking recent stories requires access to current and potentially rapidly changing knowledge.

While a knowledge base can be universal and built collaboratively from many sources, the fact-checking process is constrained to specific documents. No reliable automatic method for extracting check-worthy facts has been created yet. One possible approach towards this end has been described in Hassan et al. (2017). The authors created the ClaimBuster fact-checking platform, which contains a claim spotting model built on top of a human-labelled dataset of check-worthy claims. The system uses machine learning and natural language processing of live discourses, social media, and news to identify factual claims which are compared against a database of facts verified by professionals.

Fact confirmation or refutation based on a knowledge base requires a sophisticated search through the database. Ciampaglia et al. (2015) present a method for verifying specific claims by finding the shortest path between concept nodes in a knowledge graph of facts extracted from Wikipedia infoboxes. Published and generally known facts are often not fully covered in one specific knowledge base. Trivedi et al. (2018) describe LinkNBed, a framework able to effectively process multiple knowledge graphs and identify entity links across the databases. The linkage information then allows the resulting facts to be combined from different knowledge bases.

The fact extraction and verification (FEVER) shared task (Thorne et al. 2018) provides a comparison of 23 competing systems for automated fact-checking. The benchmark dataset consists of 185,445 human-generated claims, manually evaluated against textual evidence from Wikipedia to be true or false. The best scoring participant managed to obtain a 64.21% accuracy in correctly classified claims. The approach described in Nie et al. (2019) improves this result to obtain 66.49% accuracy.

Fact-checking is possibly the most reliable and accurate approach of detecting fake news; current automated methods serve mainly as (advanced) supporting tools. Approaches which can be deployed now must be human assisted, with annotators extracting claims from articles while fact-checking against a knowledge graph is provided automatically.

### 5.3.2 Manipulative style recognition

The assumption on which these methods are based is that the veracity of text can be assessed from its secondary characteristics and not directly from the semantics of the message itself. The mechanism has been theorised to be possibly subconscious (Zhou, Zhang 2008; Siering et al. 2016)—if the author knows that a piece of information is deceptive, or his intent is malicious, he or she will change the way the message is formulated. This theory has been confirmed in practice, and many successful methods based on this approach have been devised.

The general task is to predict, for a given article or post, whether it is deceptive or not. Older methods tend to operate on the whole investigated piece, while recent approaches are more fine-grained and also attempt to pinpoint the exact locations in the text in which deceptive techniques appear.

The methods build on the standard machinery of natural language processing and machine learning in which text classification has been studied extensively. The task is usually specified in a supervised setting. An annotated corpus consisting of representative examples of both truthful and deceptive posts is used as a training set for a machine learning classifier. The classifier attempts to find patterns in the training data, and it is then able to predict the authenticity of previously unseen articles.

In contrast to older methods which train the classifiers on various hand-crafted features extracted from the text ranging from simple measures, such as the presence of specific words or phrases, the amount of special characters, expletives, spelling errors, length of sentences, and the ratio of modal verbs, to complex ones, such as the writer's stance towards the topic discussed, its readability score, or the syntactic structure of the text, recent approaches widely employ *deep learning* methods where the classifier operates directly on the source text.

The main issue for style-based methods lies in constructing the gold standard datasets as humans have been shown to be poor at detecting deception (Rubin 2010). Nevertheless, style-based methods are a very active research area, possibly for multiple reasons: No external information is necessary, only the content itself; text classification methods are well studied in natural language processing, with many different applications; and style-based methods generalise well and can be easily applied to previously unseen data in isolation. Some interesting results of style-based deception detection are presented below.

One of the first methods is described in Burgoon et al. (2003). The authors aim to discriminate deceptive chat communications using a decision tree classifier on simple features extracted from the text. Song et al. (2012) describe their experiments in detecting deceptive reviews and essays by adding the syntactic structure of text to word sequence and part-of-speech features, and they note that syntactic features along with unigrams reach the best accuracy. Chen et al. (2015) suggest detecting misleading content by detecting *clickbait*[2] in news article headlines using support vector machines, but they do not provide a rigorous evaluation. Rubin et al. (2015) look at using features based on rhetorical structure theory (Mann, Thompson 1987) for identifying deceptive news along with a logistic regression based classifier. While the reported accuracy is low, the authors claim this might be due to the limited amount of training data. The work of Popoola (2017) evaluates rhetorical structure theory features against deceptive Amazon reviews and notes a significant correlation.

---

2 Hyperlinks or headlines crafted to deceptively attract attention.

The experiments of Rubin et al. (2016) describe a predictive method for discriminating between satirical and truthful news articles using support vector machines based on an article's vocabulary and additional features quantifying negative affect, absurdity, humour, grammar, and punctuation. The obtained precision is 0.90, recall 0.84. Reis et al. (2019) compare multiple classifiers on a large set of features extracted from the BuzzFeed dataset and conclude that XGBoost (Chen, Guestrin 2016) and Random Forests (Breiman 2001) provide the best results in the context of fake news classification. This supports the conclusions of Fernández-Delgado et al. (2014), which compares 179 different machine learning classifiers in a more general setting. While the previously described approaches use a diverse set of classifiers, the extracted features can be used to train any of the classifiers presented. Similarly Horák et al. (2019) applied a variety of classifiers to the dataset described in section 4.4.4, 'Dataset of propaganda techniques in Czech news portals (MU dataset)', achieving accuracy up to 0.96 and weighted $F_1$ 0.85 with support vector machines trained using stochastic gradient descent.

A thorough evaluation of Mitra et al. (2017) uncovers words and phrases which correlate with a high or low credibility perception of event reports on Twitter. While this information does not directly provide a signal related to fake news, it can be used to assess how credible and, therefore, dangerous a post will appear to be.

The approaches presented so far process the input texts in a limited workflow scenario. First, each text is analysed and a set of preselected features (binary or numeric) is extracted in a table row form. All automated processing then works only with the resulting table. Such an approach reveals and summarises important aspects of the input news article, which are often sufficient and necessary for the resulting decision. However, the tabular methods are not able to distinguish subtle differences in the meaning based on the order of information (e.g. words) in the text. This is the reason why recurrent neural network (RNN) architectures, such as long short-term memory (LSTM) networks (Hochreiter, Schmidhuber 1997), have been designed to operate on word sequences instead of just extracted tabular features and are able to discriminate based on long distance meaning dependencies in the input context. The capture, score, and integrate (CSI) model (Ruchansky et al. 2017) employs a complex hybrid system which uses information about user engagement and source in addition to the article text and trains a recurrent neural network with this data. The reported accuracy on a Twitter dataset for classifying false information is 0.89.

Volkova et al. (2017) evaluate various deep neural network architectures against traditional methods and find a significant improvement in classification accuracy when including certain social network and linguistic features. The presented neural network models benefited strongly from additional inputs denoted as bias cues (e.g. expressions of tentativeness and possibility as well as assertive, factive, or implicative verbs), subjectivity cues (positive and negative subjective words and opinion words), psycholinguistic cues (persuasive and biased language), and moral foundation cues (appeals to moral foundations, such as care and harm, fairness and cheating, or loyalty and betrayal). Ajao et al. (2018) describe an RNN model for the detection of fake news on Twitter and achieve 0.82 accuracy on the PHEME dataset (Zubiaga et al. 2016), beating the previous state-of-the-art result.

The shared task described in Da San Martino et al. (2019) evaluated the performance of 25 systems for detection of 18 different deceptive techniques in news articles and the specific locations in which they appear. The most successful approaches employ the BERT (bidirectional encoder representations from transformers) language model (Devlin et al. 2018) to obtain an abstract representation of the text. The best reported $F_1$ for identifying deceptive techniques at the sentence level is 0.63. The best result for obtaining the locations reaches $F_1$ 0.23. The dataset used for training and evaluation of the techniques is described in section 4.4.3, 'Dataset for fine-grained propaganda detection (QCRI dataset)'.

### 5.3.3 Source credibility

An article published on an unreliable website by an unreliable user is much more likely to be unreliable. From this perspective, the reliability of an article can be assessed independently of its content. In the analysis of Silverman (2016), it is shown that a vast majority of fake news comes from either hyper-partisan websites or fake news websites pretending to be regular news outlets. Therefore, identifying spam websites can assist in identifying unreliable sources. Traditional website reliability metrics which had been used by search engines, such as PageRank (Page et al. 1999), are not useful today as spammers have managed to overcome them, so new approaches are necessary.

The work of Esteves et al. (2018) provides a method of assessing the credibility of news sites by applying various machine learning methods to indicators, such as the article text content (e.g. text category, outbound links, contact information, or readability metrics) and the article metadata (e.g. the website domain, the time of the last update, or specific HTML tags). The authors exclude social-based features, such as popularity and link structure, as these rely on external data sources which can be easily manipulated and access to the information at scale is expensive.

Another approach attempts to identify automated malicious users and bots that spread misleading information. Mehta et al. (2007) characterise the behaviour of spam users and the patterns of their operations and propose a statistical method of identifying these users based on outlier detection.

Abbasi and Liu (2013) describe the CredRank algorithm which quantifies user credibility. The main idea is that malicious users cooperate and form larger and more coherent clusters compared to regular users, who are likely to form smaller clusters. Shu et al. (2017) devise a framework for evaluating news credibility based on the relationship between the publisher, the news piece, user engagement, and social links among the users. These metrics are obtained from different sources and then the resulting score is extracted using optimisation methods.

### 5.3.4 Information flow analysis

This approach is based on the patterns in which fake news propagates—how users and other sources interact with it, and how it is shared. As actual empirical information on the prevalence of fake news is sparse, studies in this field also commonly investigate rumours or unconfirmed news which can be identified more easily.

A central concept for this approach is a *propagation tree* or *propagation cascade* (Wu et al. 2015; Vosoughi et al. 2018). The tree consists of nodes, representing posts, and edges, which connect pairs of posts. The edges represent the relationship between the posts—commonly *is a share of* or *is a response to*. The root of the propagation tree is the original post.

The propagation patterns of fake news and those of regular news differ. Vosoughi et al. (2018) analyse the diffusion of verified true and false news stories on Twitter and report that falsehoods diffuse significantly farther, faster, and broader than truth. This effect was even more pronounced for political news. The authors report that they did not observe any acceleration in the propagation of false news due to the effect of bots, which suggests that humans, not bots, are the cause of the faster spread of false news.

The inherent weakness of these methods is their need to first observe the behaviour of a significant amount of users in order to make any judgements, so their predictive power is low in the early diffusion stages; reliable predictions can be obtained only after most of the damage had already been done.

## 5.4 Datasets and evaluation

Despite the importance of the task, there are only a few existing datasets suitable to evaluate automatic methods for the analysis and detection of propaganda. In this section, we describe them in detail.

In general, the datasets are rather small, which is due to the often complex annotation process. Annotators need to go through specific training and the annotation itself is also very tedious. Annotation schemes differ, so the datasets are hard to compare as each of them serves a different purpose and is suitable for different tasks. The datasets are also heterogenous and not all are in English.

### 5.4.1 Trusted, satire, hoax, and propaganda (TSHP) 2017 corpus

For the purpose of language analysis of fake news, Rashkin et al. (2017) have prepared a dataset comprising articles from eleven sources and labelled with four classes: trusted (articles from Gigaword News–see Graff, Cieri 2003), satire (The Onion, The Borowitz Report, Clickhole), hoax (American News, DC Gazette) and propaganda (The Natural News, Activist Report), together 22,580 news articles fairly balanced between the classes. The data is available for download (Rashkin, n.d.; Rashkin et al. 2017). The accompanied linguistic analysis showed that the level of news reliability can be predicted using the detection of certain language devices, such as subjectives (brilliant), superlatives, or action adverbs (foolishly).

### 5.4.2 QProp corpus

Barrón-Cedeño et al. (2019) have built a dataset containing 52 thousand articles from over one hundred news sources. The articles are annotated on the document level with either 'propagandistic' (positive) or 'non-propagandistic' (negative) labels. If an article comes from a source considered 'propagandistic' by Media Bias Fact Check (Media Bias/Fact Check, n.d.), then it is labelled as positive. Authors also added meta-information from the GDELT project (Global Database of Events, Language, and Tone, see Leetaru, Schrodt 2013). The corpus is available for download (Barrón-Cedeño et al. 2019).

### 5.4.3 Dataset for fine-grained propaganda detection (QCRI dataset)

The dataset has been used in the shared task on fine-grained propaganda detection, organised in 2019 as a part of the 'Conference on Empirical Methods in Natural Language Processing' and the '9th International Joint Conference on Natural Language Processing'. It has also been used for 'Hack the News Datathon Case—Propaganda Detection' held in January 2019 and in 'Semeval 2020' for Task 11. It has been provided by researchers from the Qatar Computing Research Institute (QCRI).

The authors worked with propaganda defined as 'whenever information is purposefully shaped to foster a predetermined agenda'. The propaganda in the dataset is classified into the following 18 types of manipulative techniques:

1. loaded language (strongly positive and negative, emotionally loaded vocabulary);
2. name-calling or labelling (linking subjects with words of fear, hate, desire, and other emotions);
3. repetition (i.e. 'a lie that is repeated a thousand times becomes truth');
4. exaggeration or minimisation;
5. doubt;
6. appeal to fear/prejudice;
7. flag-waving (playing on a strong national, ethnic, racial, cultural, or political feeling);
8. causal oversimplification (replacing a complex issue with one cause);
9. slogans;
10. appeal to authority;
11. black-and-white fallacy, dictatorship[3] (eliminating many options with only two alternatives or even with only a single right choice);
12. thought-terminating cliché (e.g. 'stop thinking so much' or 'the Lord works in mysterious ways');
13. whataboutism (do not argue but charge opponents with hypocrisy);

---

3 This manipulative technique is sometimes referred as a *false dilemma*.

14. *reductio ad Hitlerum* (Hitler hated chocolate, X hates chocolate, therefore X is a Nazi);

15. red herring (divert attention away by introducing something irrelevant);

16. bandwagoning (a form of *argumentum ad populum*);

17. obfuscation, intentional vagueness, and confusion (deliberately unclear language); and

18. straw man (arguing with a false and superficially similar proposition as if an argument against the proposition were an argument against the original proposition).

A more detailed explanation of the techniques is described in Da San Martino et al. (2019). The dataset has been created by a private company named A Data Pro and contains 451 articles gathered from 48 news outlets (372 from propagandistic and 79 from non-propagandistic sources). The dataset contains 21,230 sentences (350,000 words) with 7,485 instances of propaganda technique use. The most common are *loaded language* and *name-calling/labelling.*

The inter-annotator agreement has been assessed with a gamma measure (Mathet et al. 2015) suitable for tasks with annotations containing potentially overlapping spans. For an independent annotation of four annotators and six articles $\gamma = 0.31$. To improve this rather low agreement, the annotation schema was changed, and pairs of annotators came up with a final annotation together with a consolidator. This yielded a significantly higher $\gamma$ (up to 0.76) when measured between an individual annotator and the appropriate consolidated annotation.

Below is a sentence example with an annotation:

'In a glaring sign of just how [400]stupid and petty[416] things have become in Washington these days, Manchin was invited on Fox News Tuesday morning to discuss how he was one of the only Democrats in the chamber for the State of the Union speech [607]not looking as though Trump [635]killed his grandma[653].'

The three fragments are labelled as follows:
1. 400–416 loaded language;
2. 607–653 exaggeration or minimisation; and
3. 635–653 loaded language.

### 5.4.4 Dataset of propaganda techniques in Czech news portals (MU dataset)

Since 2016, researchers from the Department of Political Science at the Faculty of Social Studies, Masaryk University (MU) have been collecting and manually annotating propaganda techniques at document-level in articles from four Czech media outlets[4] with a frequent pro-Russian bias and/or manipulative content as the original research focus has been mainly on pro-Russian propaganda. Since 2017, the annotation has been made more fine-grained, and techniques have also been annotated at the phrase-level. This was accomplished by using a dedicated editor built by researchers from the Natural Language Processing Centre at the Faculty of Informatics, MU.

The dataset contains binary and multi-value attributes with the phrase-level attributes capturing the presence of a certain kind of manipulation and the document-level attributes framing the broader context of the news article.

Phrase-level attributes (with possible values) include:

1. blaming (yes/no/not sure; accusing someone of something);
2. labelling (yes/no/not sure);
3. argumentation (yes/no/not sure; does the text contains arguments for or against a proposition?);
4. emotions (outrage/compassion/fear/hatred/other/not sure);
5. demonising (yes/no/not sure; extreme form of negative labelling);
6. relativising (yes/no/not sure);
7. fearmongering (yes/no/not sure; appeal to fear, uncertainty, or threat);
8. fabulation (yes/no/not sure; rumouring and fabrication);
9. opinion (yes/no/not sure; does the text contain the clearly stated opinion of the author?);
10. location (EU/Czech Republic/USA/other country/Russia/Slovakia/not sure);
11. source (yes/no/not sure; is the proposition backed up with a reference?);
12. Russia (positive example/neutral/victim/negative example/hero/not sure; how is Russia portrayed?);
13. expert (yes/no/not sure; is the fact corroborated by an expert?); and

---

4 www.sputnik.cz, www.parlamentnilisty.cz, www.ac24.cz, and www.svetkolemnas.info.

14. attitude towards a politician (positive/neutral/negative/acclaiming/not sure).

Document-level attributes include:

15. topic (migration crisis/domestic policy/foreign policy [diplomacy]/society [social situation]/energy/social policy/conflict in Ukraine/culture/conflict in Syria/arms policy/economy [finance]/conspiracy/other);

16. genre (news/interview/commentary);

17. focus (foreign/domestic/both/not sure); and

18. overall sentiment (positive/neutral/negative).

The dataset was described in detail in Horák et al. (2019) but was later enlarged with annotated data from 2018. It contains 5,500 documents from 2016, 1,994 documents from 2017, and 2,200 documents from 2018. The documents from 2016 are annotated one annotator per article, but before annotating there was a pilot phase in which the annotators were trained and tested including multiple-round control of the inter-annotator agreement[5]. The other documents have been annotated by three annotators, so the inter-annotator agreement can be measured. If at least two annotators agreed upon a value of an attribute, it was included in the final dataset. The overall percentage agreement has been around 80%; however, as attributes differ in value sets, the average Cohen's kappa (Cohen 1960) ranges from 0.2 (relativisation) to 0.7 (location), clearly showing the difficulty of the annotation task.

An example of annotated data[6]:

Film director Nvotová: ((Slovakia)[location=Slovakia] is rotten)[labelling=yes], (its politics has its brutal roots in corruption.)[argumentation=yes] President of the (Czech Republic)[location=Czech republic] Miloš Zeman said during his inaugural speech that there were no voters in the better and worse category. ('The president should not grade political parties

5 Inter-coder reliability was tested using Cohen's kappa. In total, five rounds of pilot coding were conducted before the results for each variable were satisfactory. The most difficult was to find the annotators ability to identify the presence of the author's opinion in the text (0.63), and the manipulative technique of relativisation (0.65) was moderate, while the level of agreement on the presence of the manipulative technique of labelling (0.89) was strong (with other variables' scoring in between).

6 The dataset is in Czech. The example was translated in English by authors. The Czech original is: Režisérka Nvotová: Slovensko je prohnilé, tamní politika má brutální kořeny korupce. Prezident České republiky Miloš Zeman během svého inauguračního projevu prohlásil, že neexistují voliči první a druhé kategorie. 'Prezident by neměl známkovat politické strany, protože nejsou strany první a druhé kategorie', řekl prezident Zeman.

because there are no better and worse category parties',)[argumentation=yes] (said)[source=yes] President Zeman.

### 5.4.5 Dataset to study fake news in Portuguese

Moreno and Bressan (2019) introduced corpus FACTCK.BR, a dataset to study fake news. It contains 1,309 claims in paragraph form (a short text) which have been fact-checked by one of nine Brazilian fact-checking initiatives. Each claim consists of the following items:

1. URL of origin;
2. fact-checking author;
3. publishing date;
4. date the claim was reviewed;
5. the claim itself;
6. title of the article;
7. rating of the veracity;
8. best rating (based on various ratings); and
9. text label (various fact-checking agencies use different labels—false, true, impossible to prove, exaggerated, controversial, inaccurate, etc.).

The data items in this dataset are short texts but, in fact, the annotation is document-level. This makes this resource similar to the Proppy corpus.

### 5.4.6 Topics and emotions in the Russian propaganda dataset

The study from Miller (2019) has used a dataset consisting of roughly two hundred thousand tweets from 3,814 Twitter accounts associated by Twitter with the Russia-based Internet Research Agency (Popken 2018). The same dataset was used in the special counsel's investigation (2017–19) of Russian interference in the 2016 United States elections. The dataset does not contain manual annotation but is useful for the analysis of topics, keywords, and emotions in Russian propaganda on social media.

### 5.4.7 The BuzzFeed-Webis Fake News Corpus 2016

This dataset, introduced in Potthast et al. (2017), contains a sample of posts published on Facebook from nine news agencies close to the 2016 United States election. Posts and linked articles from mainstream, left-wing, and right-wing publishers have been fact-checked by five journalists. It contains 1,627 articles—826 mainstream, 356 left-wing, and 545 right-wing articles. Posts have been labelled as mostly true, mixture of true and false, mostly false, and no factual content if the post lacked a factual claim.

### 5.4.8 LIAR

Wang (2017) gathered 12,836 short statements labelled by fact-checkers from PolitiFact. The statements come from news releases, television or radio interviews, and campaign speeches. The labels represent a range of fact-checked truthfulness including 'pants on fire' (utterly false), false, barely true, half true, mostly true, and true.

### 5.4.9 B.S. Detector dataset

This dataset (Risdal 2016) has been collected from 244 websites classified by a browser extension B.S. Detector[7] developed for checking (and notifying users of) news truthfulness. It comprises the texts and metadata of 12,999 posts.

### 5.4.10 CREDBANK

Mitra and Gilbert (2015) crowdsourced a dataset of approximately 60 million tweets covering the end of 2015. The tweets have been linked to over a thousand news events and each event has been assessed for credibility by 30 annotators from Amazon Mechanical Turk.

---

7 B.S. here stands for bullshit.

**Table 5.1: Overview of datasets and corpora**

| Name | Data + annotation | Approx. size | Lang |
|------|-------------------|--------------|------|
| TSHP | web articles in four classes | 22,000 articles | En |
| Qprop | news articles in two classes | 52,000 articles | En |
| QCRI dataset | news articles labelled with manipulation techniques | 451 articles | En |
| MU dataset | news articles labelled with manipulation techniques | 9,500 articles | Cs |
| FACTCT.BR | statements rated by veracity | 1,300 paragraphs | Pt |
| IRA twitter | unclassified tweets | 3,800 tweets | En |
| BuzzFeed | Facebook posts | 1,600 articles | En |
| LIAR | statements labelled with truthfulness | 13,000 statements | En |
| BS detector | web pages in a few classes | 13,000 posts | En |
| CREDBANK | tweets linked to events classified by credibility | 60M tweets | En |

*Source:* Authors.

## 5.5 Summary

In this chapter, we have summarised the latest approaches to the automatic recognition and generation of fake news, disinformation, and manipulative texts in general. The technological progress in this area accelerates the dispersal of fictive texts, images, and videos at such a rate and quality that human forces cease to be sufficient. The importance of high-quality propaganda detection techniques thus increases significantly. Computer analyses allow the identification of many aspects of such information misuse based on the text style of the message, the information flow characteristics, the source credibility, or exact fact-checking. Nevertheless, final precautions always remain with the human readers themselves.

## 5.6 Bibliography

Abbasi, M.-A., & Liu, H. (2013). Measuring User Credibility in Social Media. In Greenberg, A.M., Kennedy, W.G., Nathan, N.D. (Eds.), Social Computing, Behavioral-Cultural Modeling and Prediction (pp. 441–448). Lecture Notes in Computer Science. Berlin, Heidelberg: Springer Berlin Heidelberg. https://doi.org/10.1007/978-3-642-37210-0_48.

Adair, B., Li, C., Yang, J., Yu, C. (2017). Progress Toward 'the Holy Grail': The Continued Quest to Automate Fact-Checking. Evanston: Northwestern University.

Ajao, O., Bhowmik, D., Zargari, S. (2018). Fake News Identification on Twitter with Hybrid CNN and RNN Models. Proceedings of the 9th International Conference on Social Media and Society - SMSociety '18. New York: ACM Press. https://doi.org/10.1145/3217804.3217917.

Auer, S., Bizer, C., Kobilarov, G., Lehmann, J., Cyganiak, R., Ives, Z. (2007). DBpedia: A Nucleus for a Web of Open Data. The Semantic Web, 4825, 722–735. Lecture Notes in Computer Science. Berlin, Heidelberg: Springer Berlin Heidelberg. https://doi.org/10.1007/978-3-540-76298-0_52.

Barrón-Cedeño, A., Jaradat, I., Da San Martino, G., Nakov, P. (2019). Proppy: Organizing the News Based on Their Propagandistic Content. *Information Processing & Management*, 56(5), 1849–1864. https://doi.org/10.1016/j.ipm.2019.03.005.

Bollacker, K., Evans, C., Paritosh, P., Sturge, T., Taylor, J. (2008). Freebase: A Collaboratively Created Graph Database for Structuring Human Knowledge. Proceedings of the 2008 ACM SIGMOD International Conference on Management of Data - SIGMOD '08, 1247. New York: ACM Press. https://doi.org/10.1145/1376616.1376746.

Breiman, L. (2001). Random Forests. *Machine Learning*, 45, 3-32. https://doi.org/10.1023/a:1010933404324.
Burgoon, J.K., Blair, J.P., Qin, T., Nunamaker, J.F. (2003). Detecting Deception through Linguistic Analysis. In Hsinchun, C., Miranda, R., Zeng, D.R., Demchak, C., Schroeder, J., Madhusudan, T. (Eds.), Intelligence and Security Informatics, 2665, 91–101. Lecture Notes in Computer Science. Berlin, Heidelberg: Springer Berlin Heidelberg. https://doi.org/10.1007/3-540-44853-5_7.

Caswell, D., & Anderson C.W. (2019). Computational Journalism. In Vos, T.P., Hanusch, F., Dimitrakopoulou, D., Geertsema-Sligh, M., Sehl, A. (Eds.), *The International Encyclopedia of Journalism Studies* (pp. 1–8). Wiley. https://doi.org/10.1002/9781118841570.iejs0046.

Cazalens, S., Lamarre, P., Leblay, J., Manolescu, I., Tannier, X. (2018). A Content Management Perspective on Fact-Checking. Companion of the The Web Conference 2018 on The Web Conference 2018 - WWW '18, 565–74. New York: ACM Press. https://doi.org/10.1145/3184558.3188727.

Chen, T., & Guestrin,C. (2016). XGBoost: A Scalable Tree Boosting System. Proceedings of the 22nd ACM SIGKDD International Conference on Knowledge Discovery and Data Mining - KDD '16, 785–794. New York: ACM Press. https://doi.org/10.1145/2939672.2939785.

Chen, Y., Conroy, N.J., Rubin, V.L. (2015). Misleading Online Content: Recognizing Clickbait as 'False News.' Proceedings of the 2015 ACM on Workshop on Multimodal Deception Detection - WMDD '15, 15–19. New York: ACM Press. https://doi.org/10.1145/2823465.2823467.

Ciampaglia, G.L., Shiralkar, P., Rocha, L.M., Bollen, J., Menczer, F., Flammini, A. (2015). Computational Fact Checking from Knowledge Networks. *Plos One*, 10(6): e0128193. https://doi.org/10.1371/journal.pone.0128193.

Cohen, J. (1960). A Coefficient of Agreement for Nominal Scales. *Educational and Psychological Measurement,* 20(1), 37–46. https://doi.org/10.1177/001316446002000104.

Da San Martino, G., Barrón-Cedeño, A., Nakov, P. (2019). Findings of the NLP4IF-2019 Shared Task on Fine-Grained Propaganda Detection. Proceedings of the Second Workshop on Natural Language Processing for Internet Freedom: Censorship, Disinformation, and Propaganda, 162–70. Stroudsburg: Association for Computational Linguistics. https://doi.org/10.18653/v1/D19-5024.

Da San Martino, G., Yu, S., Barrón-Cedeño, a., Petrov, R., Nakov, P. (2019). Fine-Grained Analysis of Propaganda in News Article. Proceedings of the 2019 Conference on Empirical Methods in Natural Language Processing and the 9th International Joint Conference on Natural Language Processing (EMNLP-IJCNLP), 5640–50. Stroudsburg: Association for Computational Linguistics. https://doi.org/10.18653/v1/D19-1565.

Devlin, J., Chang, M.-W., Lee, K., Toutanova, K. (2018). BERT: Pre-Training of Deep Bidirectional Transformers for Language Understanding. ArXiv. Association for Computational Linguistics.

Dong, X., Gabrilovich, E., Heitz, G., Horn, W., Lao, N., Murphy, K., Strohmann, T., Sun, S., Zhang, W. (2014). Knowledge Vault: A Web-Scale Approach to Probabilistic Knowledge Fusion. Proceedings of the 20th ACM SIGKDD International Conference on Knowledge Discovery and Data Mining - KDD '14, 601–10. New York: ACM Press. https://doi.org/10.1145/2623330.2623623.

Esteves, D., Reddy, A.J., Chawla, P., Lehmann, J. (2018). Belittling the Source: Trustworthiness Indicators to Obfuscate Fake News on the Web. Proceedings of the First Workshop on Fact Extraction and VERification (FEVER), 50–59. Stroudsburg: Association for Computational Linguistics. https://doi.org/10.18653/v1/W18-5508.

Fernández-Delgado, M., Cernadas, E., Barro, S., Amorim, D. (2014). Do We Need Hundreds of Classifiers to Solve Real World Classification Problems?. *The Journal of Machine Learning Research*, 15(1), 3133–3181.

Finn, S., Metaxas, P.T., Mustafaraj, E. (2014). Investigating Rumor Propagation with TwitterTrails. ArXiv.

Graves, L. (2016). Boundaries Not Drawn. *Journalism Studies*, June, 1–19. https://doi.org/10.1080/1461670X.2016.1196602.

Hassan, N., Arslan, F., Li, C., Tremayne, M. (2017). Toward Automated Fact-Checking: Detecting Check-Worthy Factual Claims by ClaimBuster. Proceedings of the 23rd ACM SIGKDD International Conference on Knowledge Discovery and Data Mining  - KDD '17, 1803–1812. New York: ACM Press. https://doi.org/10.1145/3097983.3098131.

Hochreiter, S.,& Schmidhuber, J. (1997). Long Short-Term Memory. *Neural Computation* 9(8), 1735–1780. https://doi.org/10.1162/neco.1997.9.8.1735.

Horák, A., Baisa, V., Herman, O. (2019). Benchmark Dataset for Propaganda Detection in Czech Newspaper Texts. Proceedings of Recent Advances in Natural Language Processing, RANLP 2019, 77–83. Varna: INCOMA Ltd.

Institute for Propaganda Analysis (1938). How to Detect Propaganda. *Bulletin of the American Association of University Professors,* 24(1), 49–55.

Kavanagh, J., & Rich, M. (2018). Truth Decay: An Initial Exploration of the Diminishing Role of Facts and Analysis in American Public Life. RAND Corporation. https://doi.org/10.7249/RR2314.
Klyne, G., Carroll, J.J., McBride, B. (2014, February 25). RDF 1.1 Concepts and Abstract Syntax. https://www.w3.org/TR/rdf11-concepts/. Accessed 1 December 2019.

Leetaru, K., & Schrodt, P.A. (2013). GDELT: Global Data on Events, Location, and Tone, 1979--2012. ISA Annual Convention, 2, 1–49.

Mann, W.C., & Thompson, S.A. (1987). Rhetorical Structure Theory: A Theory of Text Organization. University of Southern California, Information Sciences Institute.

Mathet, Y., Widlöcher, A., Métivier, J.-P. (2015). The Unified and Holistic Method Gamma (γ) for Inter-Annotator Agreement Measure and Alignment. *Computational Linguistics*, 41(3), 437–479. https://doi.org/10.1162/COLI_a_00227.

Mehta, B., Hofmann, T., Fankhauser, P. (2007). Lies and Propaganda: Detecting Spam Users in Collaborative Filtering. Proceedings of the 12th International Conference on Intelligent User Interfaces  - IUI '07, 14. New York: ACM Press. https://doi.org/10.1145/1216295.1216307.

Metaxas, P.T. (2010). Web Spam, Social Propaganda and the Evolution of Search Engine Rankings. In Cordeiro, J., Filipe, J. (Eds.), Web Information Systems and Technologies, 45, 170–182. Lecture Notes in Business Information Processing. Berlin, Heidelberg: Springer Berlin Heidelberg. https://doi.org/10.1007/978-3-642-12436-5_13.

Miller, D.T. (2019). Topics and Emotions in Russian Twitter Propaganda. *First Monday,* 24(5). https://doi.org/10.5210/fm.v24i5.9638.

Mitra, T., & Gilbert, E. (2015). Credbank: A Large-Scale Social Media Corpus with Associated Credibility Annotations. Proceedings of the Ninth International AAAI Conference on Web and Social Media. AAAI Press.

Mitra, T., Wright, G.P., Gilbert, E. (2017). A Parsimonious Language Model of Social Media Credibility across Disparate Events. Proceedings of the 2017 ACM Conference on Computer Supported Cooperative Work and Social Computing - CSCW '17, 126–145. New York: ACM Press. https://doi.org/10.1145/2998181.2998351.

Moreno, J., & Bressan, G. (2019). FACTCK.BR: A New Dataset to Study Fake News. Proceedings of the 25th Brazillian Symposium on Multimedia and the Web - WebMedia '19, 525–527. New York: ACM Press. https://doi.org/10.1145/3323503.3361698.

Neal, T., Sundararajan, K., Fatima, A., Yan, Y., Xiang, Y., Woodard, D. (2017). Surveying Stylometry Techniques and Applications. *ACM Computing Surveys,* 50(6), 1–36. https://doi.org/10.1145/3132039.

Nie, Y., Chen, H., Bansal, M. (2019). Combining Fact Extraction and Verification with Neural Semantic Matching Networks. Proceedings of the AAAI Conference on Artificial Intelligence, 33 (July), 6859–6866. https://doi.org/10.1609/aaai.v33i01.33016859.

Page, L., Brin, S., Motwani, R., Winograd, T. (1999). The PageRank Citation Ranking: Bringing Order to the Web. The PageRank Citation Ranking: Bringing Order to the Web.

Peleschyshyn, A., Holub, Z., Holub, I. (2016). Methods of Real-Time Detecting Manipulation in Online Communities. XIth International Scientific and Technical Conference Computer Sciences and Information Technologies (CSIT 2016), 15–17. IEEE. https://doi.org/10.1109/STC-CSIT.2016.7589857.

Popoola, O. (2017). Using Rhetorical Structure Theory for Detection of Fake Online Reviews. Proceedings of the 6th Workshop on Recent Advances in RST and Related Formalisms, 58–63. Stroudsburg: Association for Computational Linguistics. https://doi.org/10.18653/v1/W17-3608.

Potthast, M., Kiesel, J., Reinartz, K., Bevendorff, J., Stein, B. (2017). A Stylometric Inquiry into Hyperpartisan and Fake News. ArXiv Preprint ArXiv:1702.05638.

Radford, A., Wu, J., Child, R., Luan, D., Amodei, D., Sutskever, I. (2019). Language Models Are Unsupervised Multitask Learners. Technical report. OpenAi.

Rashkin, H., Choi, E., Jang, J. Y., Volkova, S., Choi, Y. (2017). Truth of Varying Shades: Analyzing Language in Fake News and Political Fact-Checking. Proceedings of the 2017 Conference on Empirical Methods in Natural Language Processing, 2931–2937. Stroudsburg: Association for Computational Linguistics. https://doi.org/10.18653/v1/D17-1317.

Reis, J.C.S., Correia, A., Murai, F., Veloso, A., Benevenuto, F., Cambria, E. (2019). Supervised Learning for Fake News Detection. *IEEE Intelligent Systems,* 34(2), 76–81. https://doi.org/10.1109/MIS.2019.2899143.

Rubin, V., Conroy, N., Chen, Y., Cornwell, S. (2016). Fake News or Truth? Using Satirical Cues to Detect Potentially Misleading News. Proceedings of the Second Workshop on Computational Approaches to Deception Detection, 7–17. Stroudsburg: Association for Computational Linguistics. https://doi.org/10.18653/v1/W16-0802.

Rubin, V.L. (2010). On Deception and Deception Detection: Content Analysis of Computer-Mediated Stated Beliefs. *Proceedings of the American Society for Information Science and Technology*, 47(1), 1–10. https://doi.org/10.1002/meet.14504701124.

Rubin, V.L., Conroy, N.J., Chen, Y. (2015). Towards News Verification: Deception Detection Methods for News Discourse. Hawaii International Conference on System Sciences.

Ruchansky, N., Seo, S., Liu, Y. (2017). CSI: A Hybrid Deep Model for Fake News Detection. Proceedings of the 2017 ACM on Conference on Information and Knowledge Management  - CIKM '17, 797–806. New York: ACM Press. https://doi.org/10.1145/3132847.3132877.

Shao, C., Ciampaglia, G.L., Flammini, A., Menczer, F. (2016). Hoaxy: A Platform for Tracking Online Misinformation. Proceedings of the 25th International Conference Companion on World Wide Web - WWW '16 Companion, 745–750. New York: ACM Press. https://doi.org/10.1145/2872518.2890098.

Shearer, E. (2018). News Use Across Social Media Platforms 2018. Pew Research Center.

Shu, K., Wang, S., Liu, H. (2017). Exploiting Tri-Relationship for Fake News Detection. ArXiv Preprint ArXiv:1712.07709.

Siering, M., Koch, J.-A., Deokar, A.V. (2016). Detecting Fraudulent Behavior on Crowdfunding Platforms: The Role of Linguistic and Content-Based Cues in Static and Dynamic Contexts. *Journal of Management Information Systems*, 33(2), 421–455. https://doi.org/10.1080/07421222.2016.1205930.

Silverman, C. (2016, November 16). This Analysis Shows How Viral Fake Election News Stories Outperformed Real News On Facebook. *BuzzFeed News*. https://www.buzzfeednews.com/article/craigsilverman/viral-fake-election-news-outperformed-real-news-on-facebook. Accessed 1 December 2019.

Song, F., Ritwik, B., Yejin, C. (2012). Syntactic Stylometry for Deception Detection. Proceedings of the 50th Annual Meeting of the Association for Computational Linguistics (Volume 2: Short Papers), 171–175. Jeju Island: Association for Computational Linguistics.

Thorne, J., Vlachos, A., Cocarascu, O., Christodoulopoulos, C., Mittal, A. (2018). The Fact Extraction and Verification (FEVER) Shared Task. Proceedings of the First Workshop on Fact Extraction and VERification (FEVER), 1–9. Stroudsburg: Association for Computational Linguistics. https://doi.org/10.18653/v1/W18-5501.

Trivedi, R., Sisman, B., Dong, X.L., Faloutsos, C., Ma, J., Zha, H. (2018). LinkNBed: Multi-Graph Representation Learning with Entity Linkage. Proceedings of the 56th Annual Meeting of the Association for Computational Linguistics (Volume 1: Long Papers), 252–262. Stroudsburg: Association for Computational Linguistics. https://doi.org/10.18653/v1/P18-1024.

Volkova, S., Shaffer, K., Jang, J.Y., Hodas, N. (2017). Separating Facts from Fiction: Linguistic Models to Classify Suspicious and Trusted News Posts on Twitter. Proceedings of the 55th Annual Meeting of the Association for Computational Linguistics (Volume 2: Short Papers), 647–653. Stroudsburg: Association for Computational Linguistics. https://doi.org/10.18653/v1/P17-2102.

Vosoughi, S., Roy, D., Aral, S. (2018). The Spread of True and False News Online. *Science* 359(6380), 1146–1151. https://doi.org/10.1126/science.aap9559.

Wang, W.Y. (2017). 'liar, Liar Pants on Fire': A New Benchmark Dataset for Fake News Detection. Proceedings of the 55th Annual Meeting of the Association for Computational Linguistics (Volume 2: Short Papers), 422–426. Stroudsburg: Association for Computational Linguistics. https://doi.org/10.18653/v1/P17-2067.

Wills, C.E., & Uzunoglu, D.C. (2016). What Ad Blockers Are (and Are Not) Doing. 2016 Fourth IEEE Workshop on Hot Topics in Web Systems and Technologies (HotWeb), 72–77. IEEE. https://doi.org/10.1109/HotWeb.2016.21.

Woolley, S.C., & Howard, P.N. (Eds.), (2018). *Computational Propaganda: Political Parties, Politicians, and Political Manipulation on Social Media*. Oxford: Oxford University Press.

Wu, K., Yang, S., Zhu, K.Q. (2015). False Rumors Detection on Sina Weibo by Propagation Structures. 2015 IEEE 31st International Conference on Data Engineering, 651–662. IEEE. https://doi.org/10.1109/ICDE.2015.7113322.

Zellers, R., Holtzman, A., Rashkin, H., Bisk, Y., Farhadi, A., Roesner, F., Choi, Y. (2019). Defending Against Neural Fake News. ArXiv, May.

Zhou, L., & Zhang, D. (2008). Following Linguistic Footprints. *Communications of the ACM* 51(9), 119. https://doi.org/10.1145/1378727.1389972.